\documentclass{article}
\usepackage{arxiv}
\usepackage{subcaption}
\usepackage{multirow}
\usepackage{cite}
\usepackage{url}
\usepackage{amsmath,amssymb,amsfonts}
\usepackage{algorithmic}
\usepackage[utf8]{inputenc}
\usepackage{graphicx}
\usepackage{textcomp}
\usepackage{xcolor}
\def\BibTeX{{\rm B\kern-.05em{\sc i\kern-.025em b}\kern-.08em
    T\kern-.1667em\lower.7ex\hbox{E}\kern-.125emX}}

\begin{document}

\title{Aspect Detection using Word and Char Embeddings with (Bi)LSTM and CRF\\
}


\author{
    Łukasz Augustyniak, Tomasz Kajdanowicz and Przemysław Kazienko \\
    Wrocław University of Science and Technology, 
    Wrocław, Poland \\
    lukasz.augustyniak@pwr.edu.pl, 
    tomasz.kajdanowicz@pwr.edu.pl, 
    przemyslaw.kazienko@pwr.edu.pl
    }

\maketitle

\begin{abstract}
We proposed a~new accurate aspect extraction method that makes use of both word and character-based embeddings. We have conducted experiments of various models of aspect extraction using LSTM and BiLSTM including CRF enhancement on five different pre-trained word embeddings extended with character embeddings. The results revealed that BiLSTM outperforms regular LSTM, but also word embedding coverage in train and test sets profoundly impacted aspect detection performance. Moreover, the additional CRF layer consistently improves the results across different models and text embeddings. Summing up, we obtained state-of-the-art F-score results for SemEval Restaurants (85\%) and Laptops (80\%).
\end{abstract}

\section{Introduction}

\subsection{Aspect Extraction in Sentiment Analysis}

Most of current sentiment analysis approaches detect the sentiment orientation for whole documents without information about the target entities (e.g., laptops) and their aspects (e.g., battery, screen, performance). By contrast, the aspect-based sentiment analysis identifies the aspects of given target entities and estimating the sentiment polarity for each of such aspects. Further, in the sentence about a~smartphone: \textit{'the screen is very clear but the battery life is crappy'}, the sentiment is positive for one aspect \textit{screen} but negative for another one: \textit{battery life}. 

\subsection{Motivation}

We wonder why not so many approaches from sequence tagging are used in aspect extraction. The sequence tagging techniques are widely used in Named Entity Recognition, Part-of-speech tagging or chunking tasks. Some research \cite{Lample2016, Ma2016, Akbik2018} proved they are a~good choice there, but they are also suitable for NLP productization, e.g. in neural networks used in spaCy \url{https://spacy.io/}, one of the best NLP library. There exist two valuable studies presenting LSTM-based models for aspect extraction, however, they were applied in the limited context. 
Li and Lam used only one pre-trained word embedding, which was utilized with classic LSTM and their own LSTM extension \cite{Li2017}. Surprisingly, the embeddings trained and published by them (ref. Amazon Reviews in Sec. \ref{sec:word_embedding}) performed very poor in our studies, see Sec. \ref{sec:results}.

Our goal was to propose comprehensive end-to-end aspect extraction method that uses general language text embeddings combined with advanced neural network architecture: LSTM/BiLSTM with an additional optional CRF layer. We wanted to evaluate our hypothesis that extending pre-trained word embedding with character embedding will improve not good enough aspect extraction models for languages other than English, e.g., Polish. As a~result, proposed by us method does not require any feature engineering or data pre-processing. To make an analysis more complete and general, we combined word and character embeddings and provide comprehensive comparison and ablation analysis of neural network model's performance. Our work is most similar to approaches applied in Named Entity Recognition (NER), but according to our best knowledge it has never been applied to aspect extraction. Simultaneously, various word embeddings extracted from different corpora were analyzed to evaluate their impact on final results. Hence, to test all most important pre-trained word embeddings available and miscellaneous approaches (Sec. \ref{sec:word_embedding}), we considered the following issues:
\begin{enumerate}
    \item How perform the general language word embeddings in the dedicated domains?
    \item What is the impact of word coverage between word embedding and the domain on quality of aspect extraction?
    \item Does character embedding improves the aspect extraction performance?
    \item Does statistical tests show that some models perform better in aspect extraction compared to the other ones? 
\end{enumerate}

Our main contributions are: (1) a~new method for aspect extraction making use of both word and char embedding, (2) comprehensive analysis of eight LSTM-based approaches to aspect extraction on five large pre-trained word embeddings. 

\section{Related Work}

\subsection{Aspect Extraction}

Researchers use several approaches for aspect-based sentiment analysis. From still commonly used rule-based methods (POS \cite{Augustyniak2017, Poria2016} or dependency-based \cite{Poria2014, Nguyen2015}), through standard supervised learning (e.g., SVMs \cite{alvarezlopez-EtAl:2016:SemEval, Brun2014} and CRF \cite{Toh2014, Chernyshevich2014} - all top approaches in SemEval2014 aspect extraction subtask) to deep learning-based approaches with CNN's or LSTM's. The interesting approach proposed by Ruder, Ghaffari, and Breslin \cite{Ruder2016}. They used a~hierarchical, bidirectional LSTM model to leverage both intra- and inter-sentence relations. Word embeddings are fed into a~sentence-level bidirectional LSTM which is passed into a~bidirectional review-level LSTM. Poria, Cambria, and Gelbukh \cite{Poria2016} proposed a~seven-layer convolutional neural network to tag each word in opinionated sentences as either aspect or non- aspect word. However, it is worth to mention that Poria also used hand-crafted linguistic patterns to improve their extraction accuracy. Another approach presented by He et al. \cite{He} proposes an attention-based model for unsupervised aspect extraction. The attention mechanism is used to focus more on aspect-related words while de-emphasizing aspect-irrelevant words. 

\subsection{Sequence Tagging}

One of the first approaches using sequence tagging for aspect extraction proposed Jakob and Gurevych \cite{Jakob:2010:EOT:1870658.1870759}. They used features such as token information, POS, short dependency path, word distance and information about opinionated sentences and build CRF model on the top of that. This work was extended with more hand-crafted features by Toh nad Wang \cite{Toh2014} in DLIREC system on SemEval 2014. The DLIREC system achieved the 1st place for restaurant and the 2nd for laptops. However, aspect extraction does not use sequence tagging schemes as often as this technique is used in Named Entity Recognition tasks \cite{Lample2016, Ma2016}. Lample, Ballesteros, Subramanian, Kawakami and Dyer \cite{Lample2016} proposed neural architecture based on bidirectional LSTM with a~conditional random field. Ma and Hovym \cite{Ma2016} introduced a~neural network architecture of bidirectional LSTM, CNN, and CRF. Hence, we see many approaches of sequence tagging in NER, but not to many applications of it in aspects extraction. 

\subsection{Text Embedding methods for Deep Learning}

\subsubsection{Word Embeddings}

Word embedding is a~text vectorization technique, and it transforms words in a~vocabulary to vectors of continuous real numbers. It is worth to mention that each word dimension in the embedding vector represents a~latent feature of this word. These vectors proved to encode linguistic regularities and patterns. The first and the most recognizable word embedding method is called  Word2Vec \cite{Le2014}. This neural network-based model covers two approaches: Continuous Bag-of-Words model (CBOW), and Skip-gram model (SG). The second excellent word embedding approach is Global Vector (GloVe) \cite{Pennington2014}, that is trained based on global word-word co-occurrence matrix. The third often used technique is fastText \cite{bojanowski2017enriching}. It is based on the Skip-gram model, where each word is represented as a~bag of character ngrams. This approach allows us to compute word representations for words that did not appear in the training data. Recently, researchers started to train and use sentiment oriented word embeddings \cite{Poria2016}. It was dictated due to the nature of the text, and they tried to include the opinionated nature of reviews which are not present in the ordinary texts. 

\subsubsection{Character Embeddings}
Beside word embeddings more and more approaches use char-based embeddings. This kind of embeddings has been found useful for morphologically rich languages and to handle the out-of-vocabulary (OOV) problem for tasks, e.g., in part-of-speech (POS) tagging, language modeling \cite{Ling2015}, dependency parsing \cite{Ballesteros2015} or named entity recognition \cite{Lample2016}. Zhang, Zhao, and, LeCun\cite{Zhang2015} presented one of the first approaches to sentiment analysis with char embedding using convolution networks. To best of our knowledge LSTM-based, char embeddings have not been used for sentiment analysis, especially for aspect extraction task. 

\section{Aspect Extraction Approaches using Word and Character Embedding with LSTM and CRF}

Due to best of our knowledge this is the first attempt to evaluate sequence tagging aspect extraction model using so many word embeddings extended with character embeddings. In the next sections we provide a~brief description of LSTM-based sequence tagging models with potential CRF layer, explain how to train character embedding models and provide all architectures used in all our experiments.

\subsection{Pre-trained Word Embedding}

The input layer for all tested architectures are vector representations of individual words. We used several word embeddings as we use pre-trained models in transfer learning. By this we try to mitigate the problem of training models based on the limited aspect training data. Our intuition is that aspect indication words should appear in regular contexts in large corpora. Moreover, there is big problem with the most of word embedding approaches related to the inability to handle unknown or out-of-vocabulary (OOV) words. It can be mitigated and is described in next section. 

\subsection{Character Embedding}

An important distinction of our work from most previous approaches is that we learn character-level embedding while training instead of hand-engineering prefix and suffix information about words or improving not enough large corpus used to train word embedding. Moreover, using character-level embedding can be advantageous for learning domain-specific representations. 

One the most common problem with word embeddings is related to out-of-vocabulary (OOV) words. A natural way of avoiding OOV is an extension of the embedding layer with character-level embeddings. The char embedding could represent words skipped in word vector representation. 

\begin{figure}[!ht]
\centering
\includegraphics[width=0.5\columnwidth]{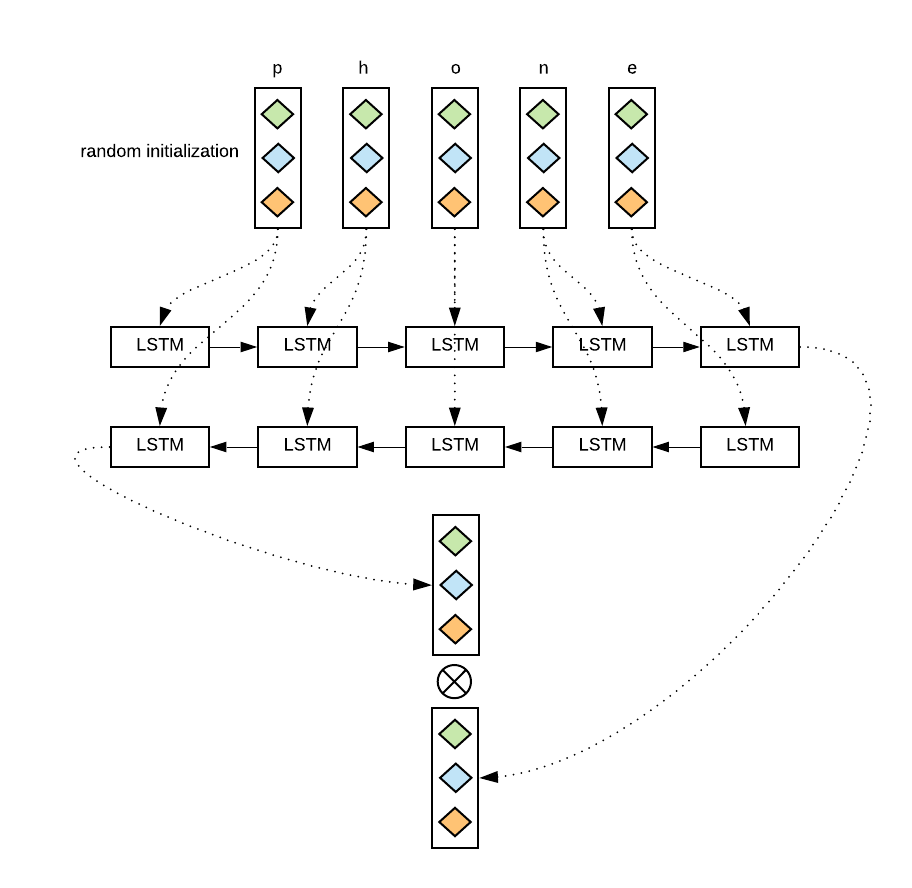}
\caption{Architecture of character embedding.\label{fig:char-embedding}}
\end{figure}

An architecture that builds word representations from individual characters processes each word separately and maps characters to character embeddings. Then, these are passed through a~bidirectional LSTM and the last states from either direction are concatenated (as in Fig.~\ref{fig:char-embedding}). The resulting vector is passed through another feed-forward layer, in order to map it to a~suitable space and change the vector size as needed. Finally, we have a~word representation, built based on individual characters.

\subsection{LSTM-based models}

LSTM networks for Long Short-Term Memory networks are a~special kind of Recurrent Neural Networks (RNN) that works great on sequential data such as sequences of words. All group of RNNs takes as an input a~sequence of vectors ($x_{1}, x_{2}, . . . , x_{n}$) and an output -- another sequence $(h_{1}, h_{2}, . . . , h_{n})$ that represents the transformed initial sequence. However, often the problem with training on sequences is how to learn the long dependencies. Hence, in the real solutions, RNNs fail in such cases and tend to be biased towards their most recent inputs in the sequence \cite{Bengio:1994:LLD:2325857.2328340}. The Long Short-term Memory Networks \cite{hochreiter1997long} have been designed to solve this issue. They are using a~memory-cell to capture exactly the long-range dependencies. LSTMs use special gates in neurons to control the proportion of the input to be put to the memory cell, and the proportion from the previous state to be forgotten. 

There are some variants of LSTMs. One is called BiLSTM \cite{DBLP:conf/icassp/2013} - bidirectional LSTM. The idea behind this architecture is to split the state neurons of a~regular RNN into two parts. The former is responsible for the positive time direction (forward states), while the latter learns the negative time direction (backward states). For sequences in the text, we feed the word or character vectors from the beginning to the end of the sentence (forward pass) and in the reverse direction (backward pass). Finally, we have two outputs from the forward and backward pass that could be concatenated into one vector representing each object in the sequence. BiLSTM enables us to train neural network faster; it decodes a~representation of a~word in the context. 

\begin{figure}[!ht]
\centering
\includegraphics[width=0.5\columnwidth]{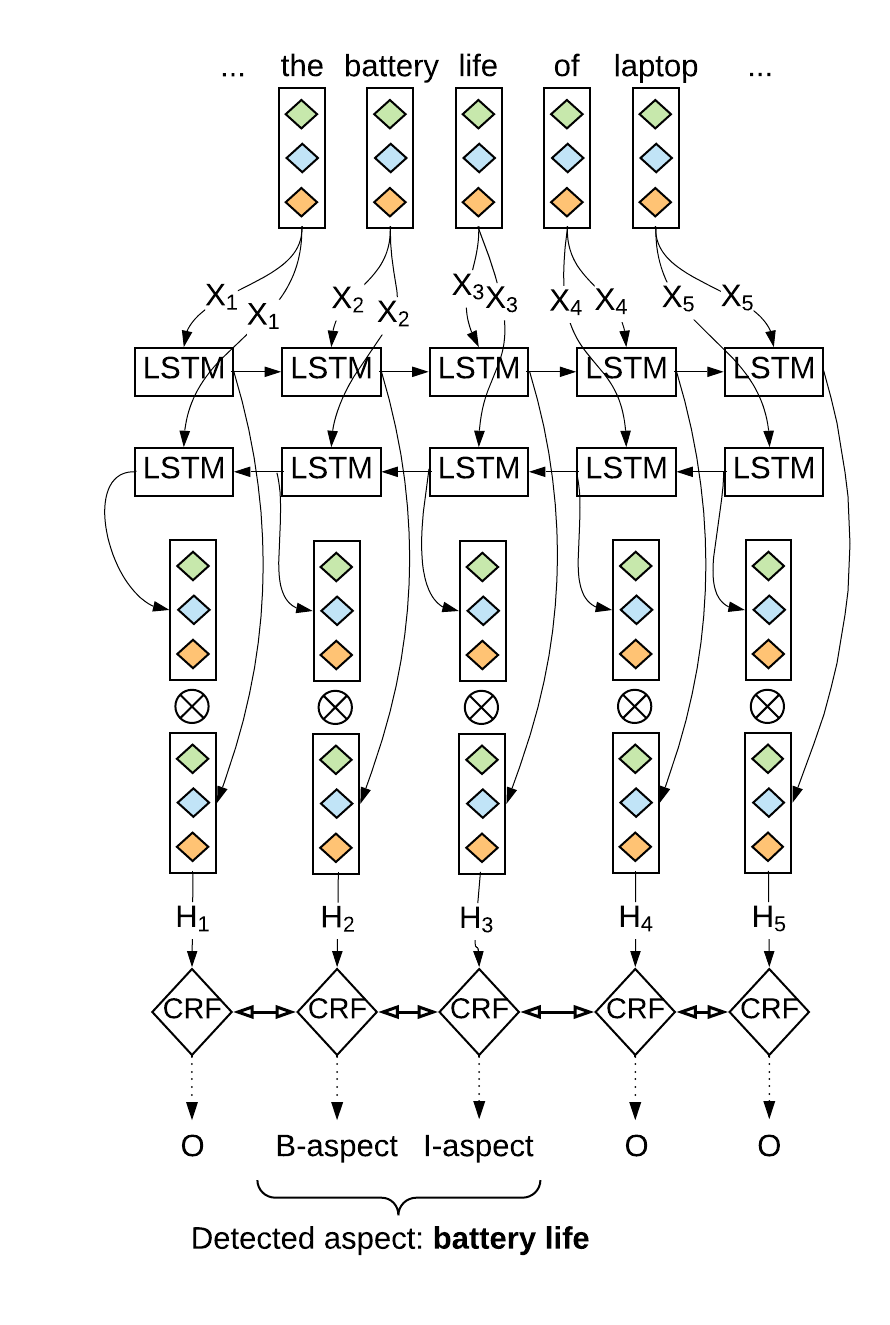}
\caption{Architecture for word embedding: BiLSTM with the CRF layer.\label{fig:word-bilstm}}
\end{figure}

\subsection{CRF layer}

The CRF layer can learn some constrains related to the final predicted labels and ensure they are valid. What is very important these constrains can be learned by the CRF layer automatically during training process. Hence, we can get constraints related to IOB scheme such as:

\begin{itemize}
    \item The label of the first word in a~sentence starts with \textit{B-aspect} or \textit{O}, but not with \textit{I-aspect};
    \item \textit{O I-aspect} is invalid sequence. The first label in aspect chunk in IOB must start with \textit{B-aspect} not \textit{I-aspect}, hence it should be replaced with \textit{O B-aspect}.
\end{itemize}

Hence, we can decrease the number of invalid predicted label sequences with such constraints. 
Using the LSTM or BiLSTM, the tagging decision is local, only some information is taken from the context. We don't make use of the neighbouring tagging predictions. For example, in \textit{battery life} with CRF we use information about tagging \textit{battery} as beginning of the aspect term and it should help us to decide that word \textit{life} should be the merge to this aspect as well. 

\subsubsection{IOB sentence coding}

We converted SemEval 2014 datasets (more in Sec. \ref{sec:datasets}) into IOB scheme \cite{Ramshaw1999}. It is a~widely used coding scheme for representing sequences. IOB is short for \textit{inside}, \textit{outside} and \textit{beginning}. The \textit{B-} prefix before a~tag (i.e., \textit{B-aspect}) indicates that the tag (\textit{aspect}) is the beginning of an annotated chunk. The \textit{I-} prefix before a~tag (i.e., \textit{I-aspect}) indicates that the tag is inside a~chunk. \textit{I-tag} could be preceded only by \textit{B-tag} or other \textit{I-tag} for ngram chunks. Finally,  the \textit{O} tag (without any tag information) indicates that a~token does not belong to any of the annotated chunks.

An exemplary sentence \textit{'I charge it at night and skip taking the cord with me because of the good battery life.'} is encoded with IOB to:
\textit{
I:O
charge:O
it:O
at:O
night:O
and:O
skip:O
taking:O
the:O
cord:B-aspect
with:O
me:O
because:O
of:O
the:O
good:O
battery:B-aspect
life:I-aspect
.:O
}

\section{Experimental setup}

\subsection{Experiment workflow}

We experimented with various sequential tagging approaches for aspect extraction. All methods are presented in Tab.~\ref{table:all_experiments}. We tested eight different configurations of features and neural networks. Moreover, we used five different pre-trained word embeddings, see Sec. \ref{sec:word_embedding} for details. In total, we evaluated 40 experimental configurations.

\begin{table}[]
\caption{All models used in our experiments. \textbf{Word} and \textbf{Char} denote the Word Embedding and Char Embedding, respectively.}
\centering
\begin{tabular}{|l|c|c|c|}
\hline
\textbf{Method abbreviation} & \textbf{Word} & \textbf{Char} & \textbf{CRF} \\ 
\hline
Wo-LSTM           & yes            & no             & no          \\
Wo-LSTM-CRF       & yes            & no             & yes        \\ 
WoCh-LSTM      & yes            & yes            & no          \\
WoCh-LSTM-CRF  & yes            & yes            & yes        \\ \hline
Wo-BiLSTM         & yes            & no             & no         \\
Wo-BiLSTM-CRF     & yes            & no             & yes       \\ 
WoCh-BiLSTM    & yes            & yes            & no         \\
WoCh-BiLSTM-CRF& yes            & yes            & yes       \\ 
\hline
\end{tabular}
\label{table:all_experiments}
\end{table}

\subsection{Text Vectorization}
\label{sec:text-vectorization}

\subsubsection{Pre-trained Word Embeddings}
\label{sec:word_embedding}

As we already mentioned, we used several different word embeddings. All of them are presented in Tab.~\ref{table:word_embeddings}. It is worth to mention that there are trained based on different sizes of corpora an with different coverage of words in language. 

\begin{table}[htbp]
\caption{All pre-trained word embeddings.}
\centering
\begin{tabular}{|l|r|r|r|}
\hline
\textbf{Word Embedding}      & \textbf{\# of words}  & \textbf{Vocab}           & \textbf{Ref} \\
\hline
Glove 840B          & 840B         & 2.2M            & \cite{Pennington14glove:global}           \\
fastText            & 600B         & 2M              & \cite{bojanowski2017enriching}            \\
word2vec            & 100B         & 3M              & \cite{Le2014}           \\
Amazon Reviews      & 4.7B         & 42K             & \cite{Poria2016}          \\
numberbatch         & 2M           & 500K            & \cite{speer-conceptnet}          \\
\hline
\end{tabular}

\label{table:word_embeddings}
\end{table}

\textbf{Glove 840B} - Global Vectors for Word Representation proposed by Stanford NLP Group, trained based on Common Crawl. \textbf{fastText} - Distributed Word Representation proposed by Facebook, trained on Common Crawl as well. \textbf{word2vec} - protoplast model of any neural word embedding proposed by Mikolov [Google] trained on Google News. \textbf{numberbatch} - Numberbatch consists of state-of-the-art semantic vectors derived from ConceptNet with additions from Glove, Mikolov's word2vec and parallel text from Open Subtitles 2016 (\url{http://opus.lingfil.uu.se/OpenSubtitles2016.php}) trained via fastText. \textbf{Amazon Reviews} - word2vec model trained on Amazon Reviews \cite{McAuley:2013:HFH:2507157.2507163}. Since it contains opinionated documents, it should be the advantage over common language texts such as Google News or Common Crawl.

\subsubsection{Char Embeddings}

We treated consecutive chars as sequences and passed it through forward and backward pass of the LSTM to get a~vector of each char that were concatenated in the end. We initialized every char vector with a~length of 25 with random numbers, hence the size of concatenated char vectors was equal to 50. We used 0.5 dropout.

\subsection{Neural Network architecture}
\label{sec:nn-architecture}

For all experiments, we used keras (\url{https://keras.io/}) with tensorflow (\url{https://www.tensorflow.org/}) and the following hyperparameters: mini-batch size: 10, max. sentence length: 30 tokens, word embedding size: 300, dropout rate: 0.5. We trained for 25 epochs using cross entropy, the Adam optimizer, and early stopping (max 2 epochs without improvement). We averaged model results according to at least 5 runs. The source code for all experiments is available at GitHub (\url{https://github.com/laugustyniak/aspect_extraction}).

\subsection{SemEval datasets}
\label{sec:datasets}

We did not use SemEval 2015 or 2016 aspect extraction datasets (in 2017 there was only aspect extraction in tweets) because they were prepared as text classification with predefined aspect categories and entities. SemEval 2014 is the last one that consists of sentences with words manually annotated as aspects.

\begin{table}[!ht]
\caption{SemEval 2014 datasets profile for Laptops and Restaurants. Multi-aspect means fraction of ngram aspects (two and more words).}
\centering
\begin{tabular}{|l|l|r|r|}
\hline
\textbf{} & \textbf{} & \textbf{Lap.} & \textbf{Rest.} \\ 
\hline
\multirow{3}{*}{Train} & \multicolumn{1}{l|}{\# of sentences} & 3,045    & 3,041        \\
                      & \multicolumn{1}{l|}{\# of aspects}   & 2,358    & 3,693        \\
                      & \multicolumn{1}{l|}{multi-aspects [\%]}   & 37    &  25       \\ \hline
\multirow{3}{*}{Test}  & \multicolumn{1}{l|}{\# of sentences} & 800     & 800         \\
                      & \multicolumn{1}{l|}{\# of aspects}   & 654     & 1,134       \\ 
                      & \multicolumn{1}{l|}{multi-aspects [\%]}   & 44    & 28        \\ \hline
\multirow{2}{*}{All}   & \multicolumn{1}{l|}{\# of sentences} & 3,845    & 3,841         \\
                      & \multicolumn{1}{l|}{\# of aspects}   & 3,012    & 4,827  \\
\hline
\end{tabular}
\label{table:semeval}
\end{table}

Both datasets are quite different. There were some issues during aspect annotation, i.e., it was unclear if a~noun or noun phrase was used as the aspect term or if it referred to the entity being reviewed as the whole \cite{Pontiki2014}. For example in \textit{This place is awesome}, the word \textit{place} most likely refers to the restaurant as the whole. Hence, it should not be tagged as an aspect term. In \textit{Cozy place and good pizza}, it probably refers to the ambience of the restaurant. In such cases, an additional review context would help to disambiguate it. Moreover, the laptop reviews often rate laptops as such without any particular aspects in mind. This domain often contains implicit aspects expressed by adjectives, e.g., \textit{expensive}, \textit{heavy}, rather than using explicit terms, e.g., \textit{cost}, \textit{weight}. We must remember that in both datasets, the annotators were instructed to tag only explicit aspects, thus, adjectives implicitly referring to aspects were discarded. The restaurant dataset contains many more aspect terms in training and in testing subsets (see Tab.~\ref{table:semeval}). The majority of the aspects in both datasets are single-words, Tab.~\ref{table:semeval}. Note that the laptop dataset consists of proportionally more multi-word aspects than the restaurants dataset. It could be one of the reasons why the average accuracy for the laptops is commonly lower than for restaurants. 

\subsection{Quality measure}

\subsubsection{F1-measure}

The F1-measure (also called F1-score or F-score) is the harmonic mean of precision and recall. It reaches its best value at 1 and worst at 0. We calculated F1-measure only for exact matches of aspects, i.e., \textit{battery life} aspect will be true positive only when both words will be tagged as aspects - no more or fewer words are possible. It is a~strong assumption opposed to some other quality measures with weak F1 when any intersection of words between annotation and prediction are treated as correctly tagged. 

\subsubsection{Nemeneyi}

Nemeneyi is a~post-hoc test used to find groups of models that differ after a~statistical test of multiple comparisons such as the Friedman test \cite{Demsar2006}. In our case, the Nemeneyi test makes a~pair-wise comparison of all model's ranks over the pre-trained word embeddings and all evaluated methods. We used alpha equal to 5\%. The Nemeneyi test provides critical distance for compared groups that are not significantly different from each other as presented in Fig.~\ref{fig:statistical-significance-analysis-restaurant-method}. 

\subsubsection{Percentage improvement}

To compare how much method $M_{2}$ improves over method $M_{1}$ we calculate the improvement according to Eq.~\ref{eq:improvement}. In other words, we show to what extent method $M_{2}$ gains within the possible margin left by method $M_{1}$, i.e., to the maximum 100\%. 

\begin{equation}
\label{eq:improvement}
    improvement(M_{1}, M_{2}) =\frac{M_{2} - M_{1}}{100\% - M_{1}}
\end{equation}

\noindent where $M_{1}$ and $M_{2}$ denote F1-measures of the first and second method, respectively. 

\subsection{Baseline Methods}

To validate the performance of our proposed models, we compare them against a~number of baselines:

\begin{itemize}
    \item \textbf{DLIREC} \cite{Toh2014}: Top-ranked CRF-based system in ATE subtask in SemEval 2014 - Restaurants domain. 
    \item \textbf{IHS R{\&}D} \cite{Chernyshevich2014}: Top-ranked system in ATE subtask in SemEval 2014 - Laptops domain.
    \item \textbf{WDEmb}: Enhanced CRF with  word  embedding,  linear context embedding and dependency path embedding \cite{Yin:2016:UWD:3060832.3061038}.
    \item \textbf{RNCRF-O} and \textbf{RNCRF-F} \cite{D16-1059}:  They used tree-structured features and recursive neural  network  as  the  CRF  input.    \textbf{RNCRF-O} was trained  without  opinion  labels. \textbf{RNCRF-F} was trained with opinion labels  and  some  additional hand-crafted  features.
    \item \textbf{DTBCSNN+F}: A convolution stacked neural network using dependency trees to capture syntactic features \cite{10.1007/978-3-319-57529-2_28}.
    \item \textbf{MIN}: LSTM-based deep multi-task learning  framework. It  jointly  handles the extraction tasks of aspects and opinions using memory interactions \cite{Li2017}.
    \item \textbf{CNN}: deep convolutional neural network using \textit{Glove.840B} word embedding as in Poria et al. \cite{Poria2016} \footnote{This approach was run by us using source code available in \url{https://github.com/soujanyaporia/aspect-extraction}.}.
\end{itemize}

The comparison of presented above models has been done in next section.

\section{Results}
\label{sec:results}

The best F1-measure were obtained by pre-trained word embeddings (\textit{Glove 840B} and \textit{fastText}) extended with char embedding using BiLSTM and CRF layer and it was 85.69\% and 80.13\% for Restaurants and Laptops respectively, see Tab. \ref{tab:results-restaurants} and Tab. \ref{tab:results-laptops}. Both models achieved performance better than the best SemEval 2014 winners (84\% and 74\%). 

\begin{table*}[]
\caption{All results averaged over 6 runs with standard deviations - Restaurant Dataset.}
\centering
\begin{tabular}{|l|l|l|l|l|l|}
    \hline
    {} &        \textbf{\scriptsize{fastText}} &  \textbf{\scriptsize{Amazon Reviews}} &     \textbf{\scriptsize{numberbatch}} &      \textbf{\scriptsize{Glove 840B}} &        \textbf{\scriptsize{word2vec}} \\
    \hline
    Wo-LSTM         &   80.8 \scriptsize{+/- 1.49} &  48.78 \scriptsize{+/- 1.02} &  76.26 \scriptsize{+/- 0.75} &   80.91 \scriptsize{+/- 1.1} &  77.73 \scriptsize{+/- 0.74} \\
    WoCh-LSTM       &  79.91 \scriptsize{+/- 1.85} &   65.81 \scriptsize{+/- 2.3} &   76.11 \scriptsize{+/- 1.9} &  81.26 \scriptsize{+/- 0.42} &  78.15 \scriptsize{+/- 0.54} \\
    Wo-LSTM-CRF     &  85.46 \scriptsize{+/- 0.21} &  52.09 \scriptsize{+/- 0.98} &  82.19 \scriptsize{+/- 0.84} &  85.02 \scriptsize{+/- 0.23} &  82.49 \scriptsize{+/- 0.32} \\
    WoCh-LSTM-CRF   &  85.25 \scriptsize{+/- 0.46} &  72.84 \scriptsize{+/- 0.62} &  \textbf{82.92 \scriptsize{+/- 0.33}} &  84.91 \scriptsize{+/- 0.38} &   \textbf{84.12 \scriptsize{+/- 0.3}} \\
    Wo-BiLSTM       &  83.17 \scriptsize{+/- 0.54} &  50.49 \scriptsize{+/- 0.87} &  78.57 \scriptsize{+/- 1.04} &  83.56 \scriptsize{+/- 0.22} &  80.16 \scriptsize{+/- 0.74} \\
    WoCh-BiLSTM     &  83.27 \scriptsize{+/- 0.61} &  69.53 \scriptsize{+/- 1.52} &  80.89 \scriptsize{+/- 0.26} &   83.55 \scriptsize{+/- 0.3} &  81.39 \scriptsize{+/- 1.08} \\
    Wo-BiLSTM-CRF   &  85.28 \scriptsize{+/- 0.46} &   50.63 \scriptsize{+/- 0.5} &  82.31 \scriptsize{+/- 0.47} &  84.96 \scriptsize{+/- 0.54} &  82.94 \scriptsize{+/- 0.51} \\
    WoCh-BiLSTM-CRF &  \textbf{85.69 \scriptsize{+/- 0.64}} &   \textbf{73.5 \scriptsize{+/- 0.91}} &  82.85 \scriptsize{+/- 0.41} &   \textbf{85.2 \scriptsize{+/- 0.28}} &  83.61 \scriptsize{+/- 1.35} \\
    \hline
\end{tabular}
\label{tab:results-restaurants}
\end{table*}

\begin{table*}[]
\caption{All results averaged over 6 runs with standard deviations - Laptops Dataset.}
\centering
\begin{tabular}{|l|l|l|l|l|l|}
    \hline
    {} &        \textbf{\scriptsize{fastText}} &  \textbf{\scriptsize{Amazon Reviews}} &     \textbf{\scriptsize{numberbatch}} &      \textbf{\scriptsize{Glove 840B}} &        \textbf{\scriptsize{word2vec}} \\
    \hline
    Wo-LSTM         &  67.75 \scriptsize{+/- 4.05} &  55.18 \scriptsize{+/- 1.77} &  57.88 \scriptsize{+/- 2.48} &  68.38 \scriptsize{+/- 3.61} &  61.59 \scriptsize{+/- 2.43} \\
    WoCh-LSTM       &  66.71 \scriptsize{+/- 4.88} &  60.01 \scriptsize{+/- 1.18} &  58.77 \scriptsize{+/- 3.86} &  70.09 \scriptsize{+/- 0.61} &   64.1 \scriptsize{+/- 2.67} \\
    Wo-LSTM-CRF     &  77.95 \scriptsize{+/- 1.79} &  65.15 \scriptsize{+/- 0.73} &   69.19 \scriptsize{+/- 2.5} &  77.72 \scriptsize{+/- 1.42} &  72.88 \scriptsize{+/- 1.12} \\
    WoCh-LSTM-CRF   &  77.53 \scriptsize{+/- 0.93} &   \textbf{70.04 \scriptsize{+/- 1.3}} &  74.15 \scriptsize{+/- 0.39} &  77.66 \scriptsize{+/- 0.46} &  75.44 \scriptsize{+/- 1.57} \\
    Wo-BiLSTM       &  73.32 \scriptsize{+/- 1.32} &  61.22 \scriptsize{+/- 1.14} &  59.02 \scriptsize{+/- 7.19} &  74.25 \scriptsize{+/- 0.87} &  67.96 \scriptsize{+/- 2.15} \\
    WoCh-BiLSTM     &  73.44 \scriptsize{+/- 2.77} &  66.06 \scriptsize{+/- 1.11} &  66.69 \scriptsize{+/- 2.07} &  73.38 \scriptsize{+/- 2.46} &  69.77 \scriptsize{+/- 2.84} \\
    Wo-BiLSTM-CRF   &  \textbf{79.34 \scriptsize{+/- 1.23}} &  64.89 \scriptsize{+/- 0.75} &  73.03 \scriptsize{+/- 1.02} &  79.99 \scriptsize{+/- 0.72} &   74.93 \scriptsize{+/- 1.0} \\
    WoCh-BiLSTM-CRF &  79.73 \scriptsize{+/- 1.36} &  69.65 \scriptsize{+/- 0.97} &  \textbf{75.09 \scriptsize{+/- 1.75}} &  \textbf{80.13 \scriptsize{+/- 0.34}} &  \textbf{76.38 \scriptsize{+/- 1.37}} \\
    \hline
\end{tabular}
\label{tab:results-laptops}
\end{table*}

\subsection{Overall Results}
\label{sec:overall-results}

We obtained the best F1-measures using \textit{Glove.840B} (80.13\% for Laptops) and \textit{fastText} (85.69\% for Restaurants) pre-trained word embeddings extended with character embedding using BiLSTM and CRF layer. Table \ref{table:overall-results} presents a~comparison of our models and baselines. It can be seen that all four of our models using either \textit{Glove.840B} or \textit{fastText} word embeddings proved to be better than any other baseline model. It is worth mentioning that our models achieved better performance than the SemEval 2014 winners - \textit{DLIREC} and \textit{IHS R\&D}. Summing up, our models' performance was superior in comparison to state-of-the-art models. 

\begin{table}[ht!]
\caption{Comparison of F1 scores on SemEval 2014.}
\label{table:overall-results}
\centering
\begin{tabular}{l|c|c|c}
\hline
\textbf{Model} & \textbf{Laptops} & \textbf{Restaurants} \\ 
\hline
DLIREC          & 73.78     & 84.01 \\
IHS R\&D        & 74.55     & 79.62 \\
RNCRF-O         & 74.52     & 82.73 \\
RNCRF-F         & \textbf{78.42}     & 84.93 \\
CNN-Glove.840B  & 77.36     & 82.76 \\
\hline
Wo-BiLSTM-CRF-fastText   & 79.34    & 85.28 \\ 
WoCh-BiLSTM-CRF-fastText & 79.73             & \textbf{85.69} \\ 
Wo-BiLSTM-CRF-Glove.840B   & 79.99            & 84.96 \\ 
WoCh-BiLSTM-CRF-Glove.840B & \textbf{80.13}            & 85.2 \\ 

\hline
\end{tabular}
\end{table}

\subsection{LSTM vs BiLSTM}

\begin{figure}[!ht]
\centerline{\includegraphics[width=0.75\columnwidth]{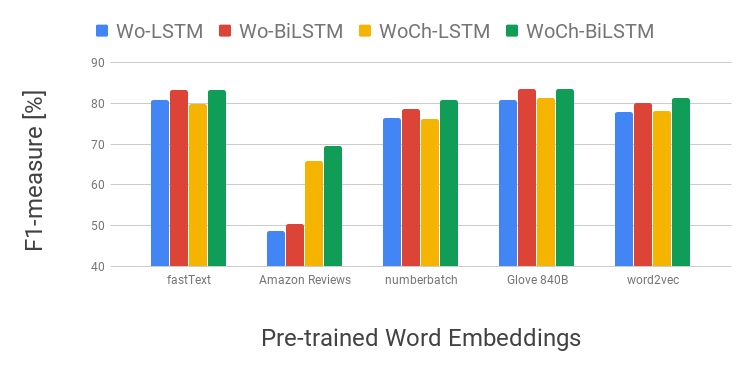}}
\caption{Comparison of LSTM and BiLSTM model's performance - Restaurant Dataset.}
\label{fig:restaurants-lstm-bilstm}
\end{figure}

We have hypothesized that BiLSTM-based model is consistently better than standard LSTM. What can be observed in  Fig.~\ref{fig:restaurants-lstm-bilstm} it was verified and proved. The figure shows a~comparison of models with LSTM and BiLSTM architecture across all evaluated pre-trained word embeddings. Interestingly, \textit{Amazon Reviews} models prove to be very poor comparing to the other embeddings, even ConceptNet-based \textit{numberbatch} and very news-based word2vec are better than \textit{Amazon Reviews} word embedding. It is even more surprising because \textit{Amazon Reviews} embedding was trained based on domain very close to laptops, i.e., Electronics.

\subsection{Influence of the CRF layer}

\begin{figure}[!ht]
\centerline{\includegraphics[width=0.75\columnwidth]{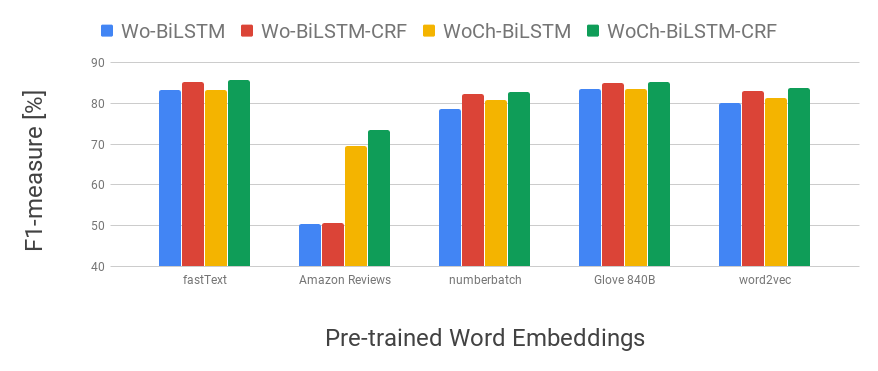}}
\caption{Analysis of CRF layer extension - Restaurant Dataset.}
\label{fig:restaurants-crf}
\end{figure}

The existence of CRF layer proves to improve the sequence tagger as well. As it can be seen at Fig.~\ref{fig:restaurants-crf} and Nemeneyi at Fig.~\ref{fig:statistical-significance-analysis-restaurant-method} models with CRF layer are most of the time significantly better than non-CRF approaches. The same pattern we saw for Laptops dataset.

\subsection{Character Embedding Extension}

\begin{figure}
\centering

\begin{subfigure}[b]{0.75\columnwidth}
    \includegraphics[width=\columnwidth]{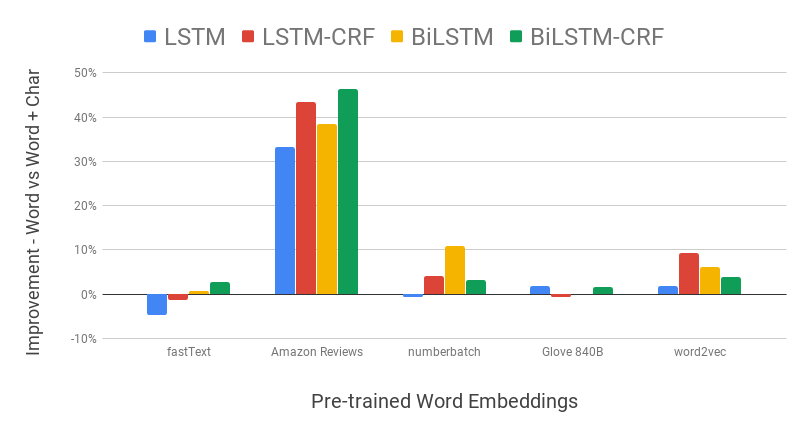}
    \caption{Restaurant Dataset.}
    \label{fig:restaurant-char-extensions}
\end{subfigure}

\begin{subfigure}[b]{0.75\columnwidth}
  \includegraphics[width=\columnwidth]{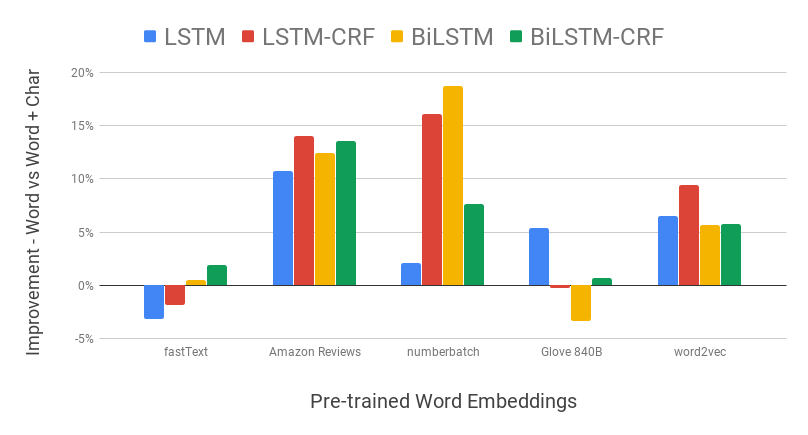}
    \caption{Laptops Dataset.}
    \label{fig:laptops-char-extensions}   
\end{subfigure}

\caption{Improvements provided by character extensions of Embedding Layer for different network architecture}
\label{fig:char-comparison}
\end{figure}

We calculated evaluated the influence of extending all neural network architectures with char embeddings according to Eq. \ref{eq:improvement}. When we analyze the word coverage between dataset (Tab. \ref{table:word_embeddings}) and used by us pre-trained word embeddings (Fig. \ref{fig:char-comparison}) we can spot that the character embeddings work very well for low coverage word embedding (\textit{Amazon Reviews} or \textit{ConceptNet numberbatch}), but it could also add noise to the good word embedding and (\textit{fastText} and Glove 840B) lower the overall performance.

\subsection{Word Embedding Vocabulary Coverage}

As we can see most of the word embeddings covers the wording of both datasets quite well. The best coverage of words contains Glove 840B and fastText models with 94\% and more coverage. The word2vec and numberbatch models present a~little less coverage between 86\% and 90\%. However, they are still good language representations. Surprisingly, Amazon Reviews model proves to be the worst in case of coverage across all embeddings with about 83\% and 67\% coverage for laptops and restaurants datasets accordingly. In the Restaurant reviews, there could appear more domain dependent words such as cousins names of ingredients, but only 83\% coverage for laptops is a~little bit unexpected, hence Amazon Reviews cover also i.e., Electronics and Laptops domains. We will investigate the influence of word embedding coverage to overall model's accuracy in next subsections. 

\begin{figure}[!ht]
\centering
\includegraphics[width=0.75\columnwidth]{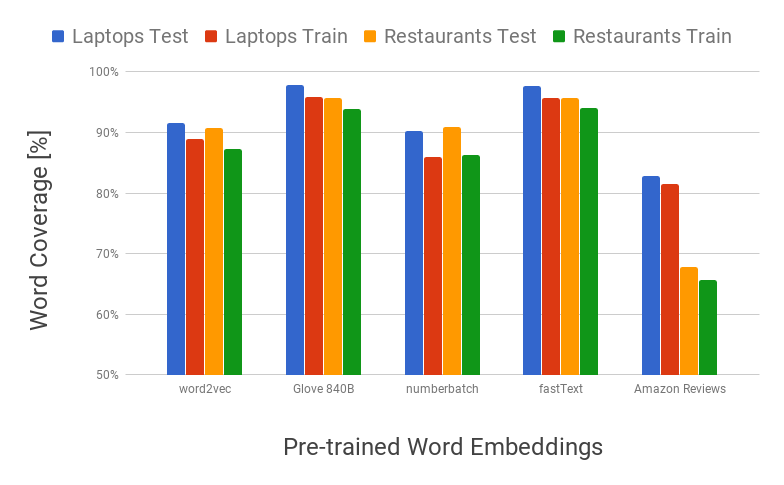}
\caption{The coverage between words in datasets and different word embeddings.}
\label{fig:words-coverage}
\end{figure}

\subsection{Statistical significance analysis}

\begin{figure}[htbp]
\centering

\begin{subfigure}[b]{0.75\columnwidth}
\centering
    \includegraphics[width=\columnwidth,height=4cm]{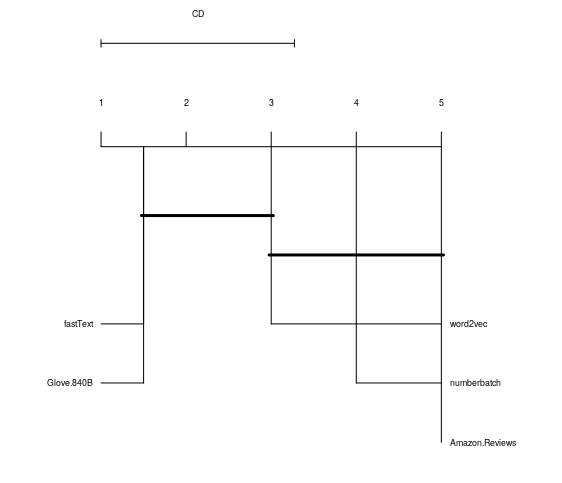}
    \caption{Different pre-trained embeddings across all evaluated methods.}
    \label{fig:statistical-significance-analysis-restaurant-embeddings}
\end{subfigure}

\begin{subfigure}[b]{0.75\columnwidth}
\centering
    \includegraphics[width=\columnwidth,height=4cm]{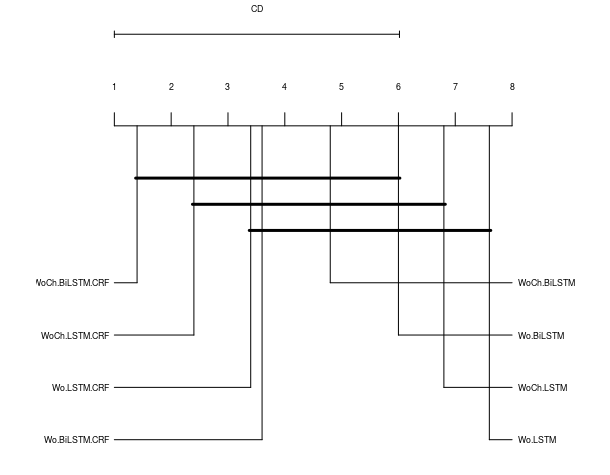}
    \caption{All evaluated methods across pre-trained word embeddings.}
    \label{fig:statistical-significance-analysis-restaurant-method}
\end{subfigure}

\caption{Nemeneyi statistical test for Restaurant Dataset.}
\label{fig:nemeneyi-restaurants}
\end{figure}

The Nemeneyi pair-wise test with Friedman rank test shows the performance compared across all pre-trained word embeddings and across all evaluated methods. As seen in Fig. \ref{fig:nemeneyi-restaurants}  \textit{Glove 840B} and \textit{fastText} word embeddings are on average the best embedding choice. Fig. \ref{fig:statistical-significance-analysis-restaurant-method} shows the significant improvements for models using CRF as the final layer. 

\section{Conclusions and Future Work}
\label{sec:conclusion-and-future-work}

We have introduced a~new accurate aspect extraction method that makes use of both word and character-based embedding. Additionally, we performed the first so wide analysis of sequence tagging approaches to aspect extraction using various neural network architectures based on LSTM and several different pre-trained embeddings.
Our method outperformed all other approaches, including the best ones from SemEval 2014 competition for both datasets available. 

We also analyzed the influence of several characteristics of word embeddings, especially out-of-vocabulary (OOV) and model settings (neural architecture type, additional CRF layer, char embedding layer) on aspect extraction performance. We proved that combining word embeddings with character-based representations makes neural architectures more powerful and enables us to achieve better, more open representations especially for models with higher OOV rates or infrequent words. It may be notably important for texts with more strongly inflected language. This opens possibilities to use such architectures with word and character embeddings for other languages such as Polish where OOV problem will be an even bigger problem due to inflected language and smaller available corpora. For that reason our future work we will focus on applying the proposed method for the Polish language. Another direction will be focused on the application of the above concepts to building complex relationships between aspects in particular hierarchies. Finally, we will use the proposed method for aspect extraction to generate abstractive summaries for various opinion datasets. 

\section*{Acknowledgment}

The work was partially supported by the National Science Centre, Poland grant No. 2016/21/N/ST6/02366, 2016/21/B/ST6/01463 and the European Union's Horizon 2020 research and innovation programme under the Marie Skłodowska-Curie grant agreement No. 691152 (RENOIR), the Polish Ministry of Science and Higher Education fund for supporting internationally co-financed projects in~2016-2019 (agreement No. 3628/H2020/2016/2) and by the Faculty of Computer Science and Management, Wrocław University of Science and Technology statutory funds.

\bibliographystyle{unsrt}
\bibliographystyle{plain}
\bibliography{ieee-aike.bib}

\end{document}